# Positive and negative explanations of uncertain reasoning in the framework of possibility theory

Henri FARRENY - Henri PRADE

Laboratoire Langages et Systèmes Informatiques
Institut de Recherche en Informatique de Toulouse
Université Paul Sabatier, 118 route de Narbonne
31062 TOULOUSE Cedex (FRANCE)

**Abstract**: This paper presents an approach for developing the explanation capabilities of rule-based expert systems managing imprecise and uncertain knowledge. The treatment of uncertainty takes place in the framework of possibility theory where the available information concerning the value of a logical or numerical variable is represented by a possibility distribution which restricts its more or less possible values. We first discuss different kinds of queries asking for explanations before focusing on the two following types : i) how, a particular possibility distribution is obtained (emphasizing the *main* reasons only) ; ii) why in a computed possibility distribution, a particular value has received a possibility degree which is so high, so low or so contrary to the expectation. The approach is based on the exploitation of equations in max-min algebra. This formalism includes the limit case of certain and precise information.

## 1 - Introduction

If we take apart pioneering works like MYCIN and TEIRESIAS (Shortliffe, 1976 ; Davis and Lenat, 1982 ; Buchanan and Shortliffe, 1984) or PROSPECTOR (see Reboh, 1983), most of the works for developing the explanation capabilities of expert systems do not take into account uncertainty. What is proposed in MYCIN is limited to the treatment of two types of questions : "how has this conclusion been established ?", and "why has the system tried to establish this fact ?". In both cases, answers are directly built by exhibiting an appropriate part of the evaluation tree of the fact to which the question refers. However, there exist various attempts to explain conclusions obtained by probabilistic inference, e.g., in the PATHFINDER system (Horvitz et al., 1986) ; see (Horvitz et al., 1988 pp. 283-286) for a survey. Moreover a new approach has been recently proposed by Strat (1987), Strat and Lowrance (1988) for the explanation of results obtained in the framework of a Shafer evidence theory-based inference system. Indeed taking into account uncertainty somewhat enriches the variety of questions which are worth considering and creates further problems. This is what is explored in the following using possibility theory (Zadeh, 1978 ; Dubois and Prade, 1988) for the modeling of imprecision and uncertainty.

Section 2 gives the necessary background on the possibilistic inference method which is used in expert system shells such as TAIGER (Farreny et al., 1986) or TOULMED (Buisson et al., 1987). In Section 3, various kinds of queries asking for explanations are introduced and discussed. Section 4 proposes a unified approach, based on the exploitation of equations in max-min algebra, to the management of i) questions asking for the main facts which lead to a(n) (uncertain) conclusion, ii) explanations of the way of improving the certainty of a conclusion, and iii) explanations giving conditions which should be satisfied in order to have a particular conclusion, different from the real one. This extends the distinction between *positive* and *negative* explanations (Rousset and Safar, 1987). In the concluding remarks, the approach is briefly compared to the one developed by Strat and Lowrance.

## 2 - Possibilistic inference

### 2.1 - Basic steps of the inference process

In possibility theory the available information about the value of a single-valued attribute a for a given item x, is represented by a possibility distribution $\pi_{a(x)}$, i.e. a mapping from the attribute domain U to [0,1], which restricts the more or less possible values of a(x) ; $\pi_{a(x)}(u)$ estimates to what extent it is possible that a(x) = u ; $\pi_{a(x)}$ is supposed to be normalized, i.e. $\sup_{u \in U} \pi_{a(x)}(u) = 1$ ; this is satisfied as soon as at least one value in U is considered as completely possible (i.e. possible at the degree 1) for a(x). The state of total ignorance about the value of a(x) is represented by $\pi_{a(x)}(u) = 1$, $\forall u \in U$. The uncertainty attached to a rule "if p then q" is represented by the possibility distribution $(\pi(q|p), \pi(\neg q|p)) \in [0,1]^2$ on the two element-set $\{q, \neg q\}$ in the context p. The normalization condition writes here $\max(\pi(q|p), \pi(\neg q|p)) = 1$. We may also have a similar information in the context $\neg p$. Note that $\pi(q|p) = 1$ *and* $\pi(\neg q|p) = 0$ means that q is certainly true in the context p, while the larger

$\pi(\neg q|p)$, the more uncertain q (when $\pi(q|p) = 1$). We consider rules of the form

if $p_1$ and... and $p_n$ then q

with $p \triangleq p_1 \wedge ... \wedge p_n$. Then the inference proceeds along five distinct steps (Dubois and Prade, 1988)

i) The inference engine estimates to what extent it is possible that the elementary condition $p_i$ is satisfied (let $\pi(p_i)$ denote this possibility) and to what extent it is possible that the condition is not satisfied ($\pi(\neg p_i)$), taking into account the available information in the factual basis ; $\pi(p_i)$ and $\pi(\neg p_i)$ are obtained through a fuzzy pattern-matching technique

$$\pi(p_i) = \sup_{u \in U} \min(\mu_{P_i}(u), \pi^{P_i}(u)) ;$$
$$\pi(\neg p_i) = \sup_{u \in U} \min(1 - \mu_{P_i}(u), \pi^{P_i}(u)) \quad (1)$$

where $\mu_{P_i}$ and $\pi^{P_i}$ are the normalized membership functions of the subsets (may be fuzzy) of U, which respectively represent the condition $p_i$ and the corresponding available information (i.e. $\pi^{P_i}$ is the possibility distribution restricting the more or less possible values of the attribute concerned by $p_i$ for the considered item). When $p_i$ is a non-vague property, thus represented by a non-fuzzy subset $P_i$, we have $\max(\pi(p_i), \pi(\neg p_i)) = 1$.

ii) These possibility degrees are aggregated (in accordance to possibility theory) in order to estimate to what extent it is possible that the whole condition p holds ($\pi(p)$), or does not hold ($\pi(\neg p)$). It is assumed that the elementary conditions are logically independent. Then we have

$$\pi(p) = \min_{i=1,n} \pi(p_i)$$
and $$\pi(\neg p) = \max_{i=1,n} \pi(\neg p_i) \quad (2)$$

In case the elementary conditions are not equally important in order to detach the conclusion, formulas (2) are generalized by

$$\pi(p) = \min_{i=1,n} \max(\pi(p_i), 1 - w_i)$$
and $$\pi(\neg p) = \max_{i=1,n} \min(\pi(\neg p_i), w_i) \quad (3)$$

where the weights of importance $w_i$ satisfy the normalization condition $\max_{i=1,n} w_i = 1$. In case of a disjunction (rather than a conjunction) of elementary conditions, we have to change min into max, max into min, and $w_i$ into $1 - w_i$ in (2) and (3).

iii) The possibility degrees $\pi(q)$ and $\pi(\neg q)$ that the conclusion is true, respectively false is obtained via a matrix product

$$\begin{bmatrix} \pi(q) \\ \pi(\neg q) \end{bmatrix} = \begin{bmatrix} \pi(q|p) & \pi(q|\neg p) \\ \pi(\neg q|p) & \pi(\neg q|\neg p) \end{bmatrix} \begin{bmatrix} \pi(p) \\ \pi(\neg p) \end{bmatrix} \quad (4)$$

where the maximum operation plays the role of the sum and the min operation the role of the product (i.e. for instance $\pi(q) = \max(\min(\pi(q|p), \pi(p)), \min(\pi(q|\neg p), \pi(\neg p)))$ ). This matrix product preserves the normalization condition.

iv) Let $\mu_Q$ be the membership function which represents the restriction expressed by q, i.e. $q \triangleq$ "b(y) is Q" where b denotes the attribute underlying q and y is the considered item (we have $\pi_{b(y)} = \mu_Q$). Then the uncertainty, modelled by $(\pi(q), \pi(\neg q))$ which pervades the conclusion, induces the new possibility distribution $\pi_{b(y)}^*$

$$\pi_{b(y)}^* = \min(\max(\mu_Q, \pi(\neg q)), \max(\mu_{\neg Q}, \pi(q))) \quad (5)$$

with $\mu_{\neg Q} = 1 - \mu_Q$ ; $\pi_{b(y)}^*$ expresses that if q is uncertain ($\pi(\neg q) > 0$ and $\pi(q) = 1$), then the values outside Q are possible at the degree $\pi(\neg q)$. When $\mu_Q$ is the membership function of an ordinary (i.e. non-fuzzy) subset, (5) can be equivalently written

$$\pi_{b(y)}^* = \max(\min(\mu_Q, \pi(q)), \min(\mu_{\neg Q}, \pi(\neg q))) \quad (6)$$

this latter expression is more convenient for performing the conjunctive combination in step v).

v) Several rules may conclude on the value of the same attribute b, then the different possibility distributions $\pi_{b(y)}^*$ obtained at step iv) have to be combined ; for instance in case of two possibility distributions $\pi_{b(y)}^*{}_1$ and $\pi_{b(y)}^*{}_2$, using (6) and applying distributivity we get

$$\min(\pi_{b(y)}^*{}_1, \pi_{b(y)}^*{}_2) =$$
$$\max(\min(\mu_{Q_1}, \mu_{Q_2}, \pi(q_1), \pi(q_2)),$$
$$\min(\mu_{Q_1}, \mu_{\neg Q_2}, \pi(q_1), \pi(\neg q_2)),$$
$$\min(\mu_{\neg Q_1}, \mu_{Q_2}, \pi(\neg q_1), \pi(q_2)),$$
$$\min(\mu_{\neg Q_1}, \mu_{\neg Q_2}, \pi(\neg q_1), \pi(\neg q_2))) \quad (7)$$

This combination is clearly associative and symmetrical, and thus can be iterated. The expression (7) is easy to interpret since it is a weighted union of mutually disjoint subsets $Q_1 \cap Q_2$, $Q_1 \cap \neg Q_2$, $\neg Q_1 \cap Q_2$ and $\neg Q_1 \cap \neg Q_2$ which cover the attribute domain.

N.B. : The max-min form (6) is permitted only if Q is an ordinary subset. However this is not a severe limitation. Indeed, consider the case of two rules if p then q, and, if p then q' where $Q \subseteq Q'$ and $\pi(\neg q|p) \geq \pi(\neg q'|p)$ since a rule should be all the more certain as its conclusion is imprecise. Then, we obtained as the result of the combination step a possibility distribution equal to

$$\max(\mu_Q, \min(\mu_{Q'}, \pi(\neg q)), \pi(\neg q')) = \max[\max(\mu_Q, \min(\mu_{Q'}, \pi(\neg q|p)), \pi(\neg q'|p)), \pi(\neg p)] \text{ when } \pi(p) = 1.$$

It indicates that several rules with the same condition part and more or less precise and uncertain (but not vague) conclusions bearing on the same attribute (the more precise the conclusion, the more uncertain), are equivalent to one rule with a vague conclusion represented by a fuzzy set (with a stair-like membership function). Thus a rule with a fuzzy conclusion can be always approximated by a collection of rules with uncertain (but non-fuzzy) conclusions.

2.2 - Example

Let us illustrate this approach on the following simple example where we have four rules, which, in a *very sketchy* and incomplete way, conclude on professions which can be recommended to people

R1 : if a person likes meeting people, then recommended professions are professor or business man or lawyer or doctor

R2 : if a person is fond of creation/invention, then recommended professions are engineer or researcher or architect

R'2 : if a person is not fond of creation/invention, then he/she cannot be an engineer nor a researcher nor an architect

R3 : if a person looks for job security and is fond of intellectual speculation, then recommended professions are professor or researcher

All these rules are pervaded with uncertainty ; they are respectively represented by the matrices

$$\begin{bmatrix} 1 & 1 \\ 0.3 & 1 \end{bmatrix} \text{ for R1, } \begin{bmatrix} 1 & 0.2 \\ 0.4 & 1 \end{bmatrix} \text{ for R2 and R'2,}$$

$$\text{and } \begin{bmatrix} 1 & 1 \\ 0.3 & 1 \end{bmatrix} \text{ for R3.}$$

The numbers mirror our beliefs in the possibility of exceptions for the various rules. Let us consider a

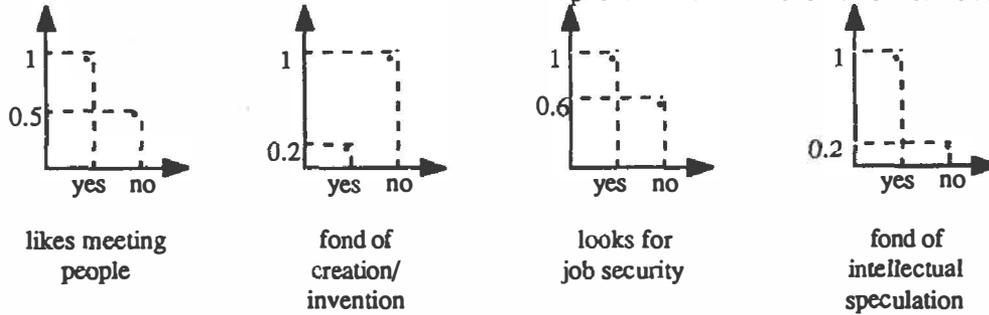

likes meeting people | fond of creation/ invention | looks for job security | fond of intellectual speculation

Figure 1

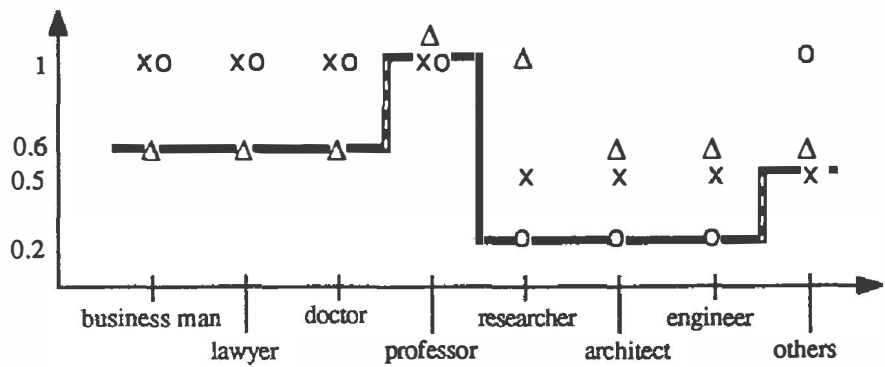

X X X : result given by R1

O O O : result given by R'2, R''2

Δ Δ Δ : result given by R3

――― : final result

Figure 2

97

person, say Peter, whose profile is indicated on Figure 1. For instance the first possibility distribution in Figure 1 indicates that it is not sure at all that Peter truly likes meeting people even if he has some propensity for that, i.e. π(Peter does not like meeting people) = 0.5. Applying the five steps of the inference procedure to this example yields the possibility distribution given in Figure 2. Note the introduction of "others" in the profession domain. It corresponds to the following understanding of the rules : professions which are not explicitly recommended (including "others") will be only possible to the extent to which the rule does not apply or has exceptions. The resulting possibility distribution may then be unnormalized for some preference profiles. A way to guarantee its normalization is to consider that "others" is implicitly among the recommended professions in any case (i.e. only professions explicitly considered by the expert can be (somewhat) excluded).

## 3 - Various kinds of queries asking for explanations

We have seen that a conclusion obtained by the inference system is represented under the form of a possibility distribution. Then the following questions can be considered

a) How is the possibility distribution obtained ? This is the direct counterpart of the MYCIN question "how". The answer is elaborated from the complete sequence of operations from which the possibility distribution results.

b) How, *mainly*, is the possibility distribution obtained ? This may be understood at least in two different ways :
- what are the main facts (and/or rules) which determine the resulting possibility distribution ? This means, for instance, that partial conclusions which are too imprecise or too uncertain to influence the final conclusion have not to be explained ;
- what are the intermediary results which may be the most surprising ones for the person who questions the system. This requires to maintain a model of user's beliefs. The extent to which a conclusion appears surprising can be computed as a degree of incompatibility between the obtained conclusion and the user's belief, i.e. by the quantity

$$1 - \sup_u \min(\mu_{Co}(u), \mu_{Be}(u))$$

where Co and Be are the fuzzy sets of possible values which represent the conclusion and the user's belief respectively. This quantity estimates to what extent the intersection between the fuzzy sets Co and Be is empty.

c) Why particular values of the conclusion domain have a possibility degree which is zero (or which is so low) ? Then the problem is to point out the key facts which determine the considered possibility degrees. More generally, one may ask why a particular possibility distribution has not been derived in place of the obtained one. We shall see in section 4 that this leads to the resolution of relational equations in max-min algebra.

d) Why the obtained conclusion is so uncertain and imprecise (i.e. all the values in the domain have a possibility degree equal or close). This may be due to i) the uncertainty and imprecision pervading the available information ; ii) to the existence of a conflict between partial conclusions which makes impossible the emergence of some value(s) ; iii) to the limitations of the inference system (e.g. many rule-based inference systems are not able to conclude "s or t" from "if p then s", "if q then t" and "p or q" ; for avoiding this limitation, the system has to be able to combine the rules themselves rather than the obtained conclusions). Note also that the research of way to improve a too uncertain or imprecise conclusion may oblige the system to not limit the investigation to the evaluation tree of the conclusion. For instance if we have two rules, one "if p then r" which is quite uncertain and another one "if q then r" which is not pervaded with uncertainty, but which was not fired due to the lack of information about q, the way to improve the conclusion is to obtain information about q.

e) How the possibility distribution of the conclusion would be modified if the possibility distribution(s) attached to (a) particular fact(s) is (are) modified ? This requires a sensitivity analysis based on analytical expressions.

f) Why did the system try to evaluate a particular fact ? This is the counterpart of the MYCIN question "why", which can be dealt with in a similar way.

## 4 - Presentation of the proposed approach

In this section we focus our attention on the relationship between facts and obtained conclusions. We are more particularly interested in the treatment of questions of type b, c, d, e considered in the preceding section. One sequence of steps i, ii, iii, iv, v (presented in section 2) constitutes an inference layer. In the course of a reasoning this process is iterated. In the following we consider the production of explanations for a given inference layer. Clearly more complex explanations can be built by iteration.

For performing explanations at the level of an inference layer, it is possible to take advantage of a system of equations (linear in max-min algebra) which relates the degrees of compatibility between facts and conditions of rules to the possibility distribution representing the conclusion. Indeed for each (uncertain) rule $R_i$ of the form "if $p_i$ then $x \in E_i$", the propagation law can be written

$$\begin{bmatrix} \pi(E_i) \\ \pi(\neg E_i) \end{bmatrix} \begin{bmatrix} \alpha_i \\ \beta_i \end{bmatrix} =$$

$$\begin{bmatrix} \pi(E_i|p_i) & \pi(E_i|\neg p_i) \\ \pi(\neg E_i|p_i) & \pi(\neg E_i|\neg p_i) \end{bmatrix} \begin{bmatrix} \pi(p_i) \\ \pi(\neg p_i) \end{bmatrix} \triangleq$$

$$\begin{bmatrix} 1 & s_i \\ r_i & 1 \end{bmatrix} \begin{bmatrix} \lambda_i \\ \rho_i \end{bmatrix}$$

i.e. $\alpha_i = \max(\lambda_i, \min(s_i, \rho_i)) = \max(\lambda_i, s_i)$
   since $\max(\lambda_i, \rho_i) = 1$
$\beta_i = \max(\min(r_i, \lambda_i), \rho_i) = \max(r_i, \rho_i)$.

The possibility distribution $\pi$ resulting of the combination of the obtained results from each rule $R_i$ is given by (in the case of 3 partial conclusions as in our example)

$$\pi = \min_{i=1,3} \max[\min(\mu_{E_i}, \alpha_i), \min(\mu_{\neg E_i}, \beta_i)]$$

$$= \max_{\substack{j=1,2;k=1,2 \\ \ell=1,2}} \min(\mu_{F_j}, \mu_{F_k}, \mu_{F_\ell}, \gamma_j, \gamma_k, \gamma_\ell)$$

$$= \max_{\substack{j=1,2;k=1,2 \\ \ell=1,2}} \min(\mu_{F_j \cap F_k \cap F_\ell}, \gamma_j, \gamma_k, \gamma_\ell)$$

with $F_j = E_1$ and $\gamma_j = \alpha_1$ if $j = 1$; $F_j = \neg E_1$
and $\gamma_j = \beta_1$ if $j = 2$; $F_k = E_2$ and $\gamma_k = \alpha_2$
if $k = 1$; $F_k = \neg E_2$ and $\gamma_k = \beta_2$ if $k = 2$;
$F_\ell = E_3$ and $\gamma_l = \alpha_3$ if $\ell = 1$; $F_\ell = \neg E_3$
and $\gamma_l = \beta_3$ if $\ell = 2$.

In our example, $E_1 \cap E_2 = \emptyset$ et $E_3 \subseteq E_1 \cup E_2$. This induces the partition $E_1 \cap \neg E_3 = \{$business man, lawyer, doctor$\}$, $E_1 \cap E_3 = \{$professor$\}$, $E_2 \cap E_3 = \{$researcher$\}$, $E_2 \cap \neg E_3 = \{$engineeer, architect$\}$, $\neg E_1 \cap \neg E_2 = \{$others$\}$. It yields

if $u \in E_1 \cap \neg E_3$  $\pi(u) = \min(\alpha_1, \beta_2, \beta_3)$
if $u \in E_1 \cap E_3$  $\pi(u) = \min(\alpha_1, \beta_2, \alpha_3)$
if $u \in E_2 \cap E_3$  $\pi(u) = \min(\beta_1, \alpha_2, \alpha_3)$
if $u \in E_2 \cap \neg E_3$  $\pi(u) = \min(\beta_1, \alpha_2, \beta_3)$
if $u \in \neg E_1 \cap \neg E_2$  $\pi(u) = \min(\beta_1, \beta_2, \beta_3)$

and finally replacing the $\alpha_i$ and $\beta_j$ by their values, it gives, in a matrix form

$$\begin{bmatrix} x_{E_1 \cap \neg E_3} \\ x_{E_1 \cap E_3} \\ x_{E_2 \cap E_3} \\ x_{E_2 \cap \neg E_3} \\ x_{\neg E_1 \cap \neg E_2} \end{bmatrix} =$$

$$\begin{bmatrix} s_1 & 1 & 1 & r_2 & 1 & r_3 \\ s_1 & 1 & 1 & r_2 & s_3 & 1 \\ 1 & r_1 & s_2 & 1 & s_3 & 1 \\ 1 & r_1 & s_2 & 1 & 1 & r_3 \\ 1 & r_1 & 1 & r_2 & 1 & r_3 \end{bmatrix} \blacksquare \begin{bmatrix} \lambda_1 \\ \rho_1 \\ \lambda_2 \\ \rho_2 \\ \lambda_3 \\ \rho_3 \end{bmatrix}$$

where ■ denotes a *min-max* product. In the example we have for the rules $r_1 = 0.3$ ; $s_1 = 1$ ; $r_2 = 0.4$ ; $s_2 = 0.2$ ; $r_3 = 0.3$ ; $s_3 = 1$, which gives

$x_{E_1 \cap \neg E_3} = x_{\text{business man or lawyer or doctor}} =$
   $\min(\max(\rho_2, 0.4), \max(\rho_3, 0.3))$  ($\alpha$)

$x_{E_1 \cap E_3} = x_{\text{professor}} = \max(\rho_2, 0.4)$  ($\beta$)

$x_{E_2 \cap E_3} = x_{\text{researcher}} =$
   $\min(\max(\rho_1, 0.3), \max(\lambda_2, 0.2))$  ($\gamma$)

$x_{E_2 \cap \neg E_3} = x_{\text{engineer or architect}} =$
   $\min(\max(\rho_1, 0.3), \max(\lambda_2, 0.2), \max(\rho_3, 0.3))$  ($\delta$)

$x_{\neg E_1 \cap \neg E_2} = x_{\text{others}} =$
   $\min(\max(\rho_1, 0.3), \max(\rho_2, 0.4), \max(\rho_3, 0.3))$  ($\epsilon$)

where $\lambda_1$ (resp. $\rho_1$) is the possibility that the person likes (resp. does not like) meeting people, $\lambda_2$ (resp. $\rho_2$) is the possibility that the person is fond of (resp. is not fond of) creation/invention, $\lambda_3$ (resp. $\rho_3$) is the possibility that the person looks for job security *and* is fond of intellectual speculation (resp. does not look for job security *or* is not fond of intellectual speculation).

Generally speaking, an equation system, such as S, relates an Input Vector IV representing the compatibility between facts and conditions of rules, an Output Vector OV describing the resulting possibility distribution and a matrix MR which caracterizes the set of rules. It can be formally written OV = MR ■ IV. There are two main ways for using this equation for explanation purposes according to what is the unknown, IV or OV.

Considering the equations of system S we can directly read the components of IV which contribute to the expression of a particular component of OV. It should be clear that any value in the domain is *a priori* completely possible and that partial conclusions may only contribute to make a particular value (more or less) impossible. For instance, equation ($\alpha$) in the example, expresses that business man, lawyer or doctor are somewhat impossible (however the possibility cannot go below 0.3 in any case) among the recommended professions if it is (quite) certain that the person is fond of creation/invention ($\rho_2$ is zero or low), or he/she looks for job security and is fond of intellectual speculation ($\rho_3$ is zero or low) ; in fact these two situations lead to other recommended professions. Obviously the expressions ($\alpha$)-($\epsilon$) enable the system to perform a straightforward sensitivity analysis of OV in terms of IV.

Let us suppose that IV is known. For instance in Peter's case $\rho_2 = 1$ and $\rho_3 = 0.6$. Then equation ($\alpha$) makes clear (since $\min(\max(1,0.4), \max(0.6,0.3)) = \rho_3 = 0.6$) that the conclusion that business man, lawyer or doctor is only possible at the degree 0.6 is mainly due to the fact that it is somewhat certain (at the degree $1 - \rho_3 = 0.4$) that the person looks for job security and is fond of intellectual speculation. Thus it is possible to point out the state of facts which determines any particular possibility degree.

Let us suppose now that OV is fixed (and IV unknown). For instance looking at Peter's case, we are astonished that his degree of possibility for having the researcher profession recommended is so low (0.2), and we ask for what would make it possible (at least) at the degree 0.8. The equation ($\gamma$) which makes obvious that the possibility degree may take any value between 0.2 and 1, leads to the condition $\rho_1 \geq 0.8$ and $\lambda_2 \geq 0.8$ ; i.e. it should be possible at least at the degree 0.8 that the person does not like meeting people (otherwise other professions would be recommended according to rule R1) and it should be possible at least at the degree 0.8 that the person is fond of creation/invention (since it is a somewhat *necessary* condition for being researcher according to R'2). Note that any pair $(\rho_1, \lambda_2) \in [0.8,1]^2$ guarantees the required degree of possibility. We see that the information conveyed by the equations of system (S), here the equation ($\gamma$), is not trivial in the sense that it encompasses the effect of several rules.

More generally, the possibility that a particular OV is obtainable can be discussed by solving the equation OV = MR ■ IV. This can be easily done using results on fuzzy relation equations (see Sanchez (1977) in particular). These results give i) the conditions of existence of a solution, ii) the expression of the smallest solution (in the sense of fuzzy set inclusion : $F \subseteq G \Leftrightarrow \mu_F \leq \mu_G$) if there exists a solution, iii) the largest solutions (there may be several), if there exists a solution.

We have only discussed the system of equations relating the compatibility of facts with conditions of rules and the possibility distribution representing the conclusion, for the sake of brevity. Two other families of max-min or min-max equations can be exploited in an explanation process, namely the expressions of the compatibility of an elementary condition of a rule with available information (given by equations (1)), and the expressions of the global compatibilities in terms of elementary ones (taking into account the level of importance of elementary conditions if they are unequal) in case of compound conditions (see equation (3)). We have presented the explanation process for one layer of inference only, for the sake of brevity. We believe that we have often to remain at this level in order to produce explanations understandable by the user. Obviously we can work across several layers of inference, by iterating the explanation process in cascade. See Farreny and Prade (1989) for details. Alternatively it would be possible to take advantage of the properties of the max-min algebra for producing out the equations relating an input vector and an output vector through several layers of rules.

Remark : In our approach what is computed by the system is a possibility distribution, which can be explained and justified to the user as shown above. However the user may also ask his/her questions in terms of *certainty* (rather than possibility). For instance, in our example, why it is not more certain that Peter should be professor ? This is equivalent to explain why it is still somewhat possible that he considers other professions, i.e. to explain the other relatively high degrees of possibility.

### Concluding remarks

In this paper we use the framework of possibility theory for representing uncertainty and imprecision. It turns out that this framework seems convenient to produce rather sophisticated explanations in a rather manageable way. The approach takes advantage of the properties of max-min algebra. It enables us also to identify easily in a min (or max) combination what are the component(s) which determine(s) the value of the result (it contrasts with operations used in other calculus like probabilities : in a product for instance, all the factors "contribute" to the result except is one of them is the neutral element '1'). In our approach we have chosen to build the explanation in terms of facts, considering that the levels of uncertainty of the rules could not be discussed and also that the expert rules were more or less known by the user. If it is not the case, the equations on which our approach is based can also be exploited for pointing out the role plaid by the uncertainty of rules.

As already said Strat and Lowrance (1987, 1988) have recently developed an approach in the framework of Shafer (1976)'s evidence theory. However in their system GISTER the uncertainty pervades only the facts (since the so-called compatibility relations they allow between universes of discourse are equivalent to imprecise but certain rules). Their system is able to explain the degree of belief of a fact and what are the facts which make not too imprecise or uncertain a conclusion. Their sensitivity analysis is based on the way a conclusion is modified when facts are discounted and makes use of entropy-like measures which respectively estimate the dissonance and the specificity (i.e. precision) of a basic probability assignment representing a fact. Mathematically speaking, possibility measures are particular cases of Shafer's plausibility functions. So, it would be possible to apply Strat and Lowrance's approach in the possibilistic case and to take advantage of specificity measures for characterising the quality of conclusions in a *global* way (dissonance measures are always zero in possibility theory ; see (Dubois and Prade, 1987) for a survey of entropy-like measures in possibility and evidence theories). The simplest measure of imprecision (or non-specificity) of a finite fuzzy set F is its scalar cardinality $|F| = \sum_u \mu_F(u)$. Clearly the smaller the cardinality of a possibility distribution the more precise the result that this distribution represents. However it is perhaps more useful to analyse the *local* influence of facts on the possibility of a particular output value, rather than their global influence on the whole possibility distribution.